\documentclass[3p]{elsarticle}

\usepackage{todonotes}
\usepackage{xcolor}
\usepackage{subcaption}
\usepackage{comment}
\usepackage{booktabs}
\usepackage{float}
\usepackage{array}
\newcolumntype{P}[1]{>{\centering\arraybackslash}p{#1}}
\usepackage{tabularx}
\usepackage{latexsym}
\usepackage{longtable}
\usepackage{latexsym}
\usepackage{graphicx}
\usepackage[
  separate-uncertainty=true,
  table-align-uncertainty=true,
]{siunitx}
\usepackage{amssymb}
\usepackage{amsmath}
\usepackage{bm}
\usepackage{rotating}

\definecolor{orange}{RGB}{255,169,64}

\newcommand{\TBD}[1]{\fbox{\bf ?}\todo[color=yellow]{{\footnotesize #1}}}
\newcommand{\nointitem}[1]{\noindent \textbf{#1}}

\journal{Biomedical Signal Processing and Control}

\begin{document}

\begin{frontmatter}



\title{On the effectiveness of smartphone IMU sensors and Deep Learning in the detection of cardiorespiratory conditions}


\author[di]{Lorenzo Simone\corref{1}}
\ead{lorenzo.simone@di.unipi.it}
\cortext[1]{Corresponding author.}
\author[di]{Luca Miglior}
\author[di]{Vincenzo Gervasi}
\author[di]{Luca Moroni}
\affiliation[di]{organization={Department of Computer Science, University of Pisa},
            city={Pisa},
            country={Italy}}

\author[ftgm]{Emanuele Vignali}
\author[ftgm]{Emanuele Gasparotti}
\author[ftgm]{Simona Celi}
\affiliation[ftgm]{organization={Fondazione Toscana Gabriele Monasterio},
            city={Massa},
            country={Italy}}

\begin{abstract}
This research introduces an innovative method for the early screening of cardiorespiratory diseases based on an acquisition protocol, which leverages commodity smartphone's Inertial Measurement Units (IMUs) and deep learning techniques. We collected, in a clinical setting, a dataset featuring recordings of breathing kinematics obtained by accelerometer and gyroscope readings from five distinct body regions. We propose an end-to-end deep learning pipeline for early cardiorespiratory disease screening, incorporating a preprocessing step segmenting the data into individual breathing cycles, and a recurrent bidirectional module capturing features from diverse body regions. We employed Leave-one-out-cross-validation with Bayesian optimization for hyperparameter tuning and model selection. The experimental results consistently demonstrated the superior performance of a bidirectional Long-Short Term Memory (Bi-LSTM) as a feature encoder architecture, yielding an average sensitivity of 0.81 ± 0.02, specificity of 0.82 ± 0.05, F1 score of 0.81 ± 0.02, and accuracy of 80.2\% ± 3.9 across diverse seed variations. We also assessed generalization capabilities on a skewed distribution, comprising exclusively healthy patients not used in training, revealing a true negative rate of 74.8\% ± 4.5. The sustained accuracy of predictions over time during breathing cycles within a single patient underscores the efficacy of the preprocessing strategy, highlighting the model's ability to discern significant patterns throughout distinct phases of the respiratory cycle. This investigation underscores the potential usefulness of widely available smartphones as devices for timely cardiorespiratory disease screening in the general population, in at-home settings, offering crucial assistance to public health efforts (especially during a pandemic outbreaks, such as the recent COVID-19).

\end{abstract}



\end{frontmatter}



\section{Introduction}


Respiratory diseases encompass a wide range of acute and chronic health conditions affecting the lung's airways and related structures, constituting a significant global health concern in terms of morbidity and mortality \cite{rdprevalence}. Tragically, an estimated 4 million people die prematurely each year due to chronic respiratory diseases. Within the category of non-communicable diseases (NCDs), asthma, chronic obstructive pulmonary disease (COPD), and occupational lung diseases are among the most prevalent chronic respiratory conditions. The latter have become major contributors to the burden of public health issues worldwide, emphasizing the need for innovative solutions.

In the context of the recent COVID-19 pandemic, the care landscape for individuals with chronic conditions has undergone significant shifts. Reduced in-person assistance contributed to the adoption of telemedicine tools that emphasize the importance of virtual-care screening, especially in low-resource clinical settings \cite{pandemic}. During the pandemic, the shortages of medications and lifestyle disruptions contributed significantly to the difficulty of managing illnesses at early stages. As a direct consequence, the implementation of affordable remote screening tools aiming at an early detection of cardiovascular diseases with high accessibility and time-cost efficiency is crucial \cite{remoteefficiency}. 
A disease-oriented definition of telemedicine has been proposed by the World Health Organization, pointing out its primary purpose of promoting healthcare access and health through remote services \cite{who_remote}. Specifically, these services are crucial in situations where distance or resources are significant factors. Within this context, novel remote monitoring technologies emerge as promising avenues, providing early assessment tools to clinicians \cite{remote} and personal monitoring systems for patients \cite{monitoring}.
The main objective of this study consists in leveraging smartphone sensors and deep learning (DL) to collect and interpret data on the kinematics of breathing, to provide an early screening tool that is usable in an at-home environment. With these objectives in mind, in this research, we propose a remote framework intended to complement physicians' efforts in the early screening of patients suffering from cardiovascular diseases. It is important to emphasize that our application is not intended to replace medical professional examinations but rather to offer valuable support and enhance the overall care experience for both patients and physicians.

Specifically, the study focuses on the acquisition of time series data from sensors recording subtle movements from the expansion and contraction of the rib cage during breathing. In particular, the proposed model analyzes temporal data recorded through accelerometer and gyroscope sensors found on widely available smartphone devices. 

We summarize the key contributions of this study as follows:
\begin{itemize}
  \item We collect a novel dataset consisting of approximately 2000 time series on more than a hundred patients in a controlled clinical setting for an early screening of cardiovascular diseases. 
  \item We propose a tailored deep learning architecture for temporal data performing supervised classification of subjects, both healthy and patients suffering from a range of cardiovascular diseases.
  \item We develop and make available a mobile application designed for autonomous data acquisition that adheres to a predefined protocol established in collaboration with medical experts.
\end{itemize}


The structure of the paper is as follows. In Section~\ref{sec:related}, we review two related research areas, the first involving sensors for measuring breathing characteristics and the other using DL models for cardiorespiratory diseases detection, comparing existing works to the present contribution. Then, we describe the data acquisition protocol (Section~\ref{subsec:data_acquisition}) examining in detail the resulting dataset and the clinical features of the population enrolled in the study (Section~\ref{subsec:population}). In subsequent subsections, we explore the proposed methodology, elucidating both the suggested preprocessing pipeline (Section~\ref{subsec:preprocessing}) and the employed deep learning architecture (Section~\ref{subsec:dl_architecture}), which underwent an extensive model selection phase yielding the experimental results reported in the following Section~\ref{sec:results}. The final segments are structured to discuss the study's outcomes and implications from a clinical perspective (Section~\ref{sec:discussion}), assessing their practical relevance followed by a critical examination of the limitations inherent in our study, acknowledging constraints and potential areas for refinement in future research. Finally, in Section~\ref{sec:conclusions} we summarize the key outcomes, providing a concise overview of the broader significance of our research and potential avenues for future work.

\section{Related work}
\label{sec:related}
To the best of our knowledge, there are no studies in the literature matching our specific research objectives and exploiting the same signal acquisition modalities. Several previous studies used similar sensing technology, focusing on inertial measurement units, such as gyroscopes and accelerometers, for detecting respiratory conditions or cardiac features. In the following we will first review several studies that confirm the reliability and feasibility of retrieving respiratory kinematics information through the placement of dynamic, remote, and wearable sensors. Then, we will examine in detail which of these approaches have included machine learning solutions as a diagnostic aid for sleep apnea or respiratory patterns detection and a wide range of digital health challenges.

\medskip

\noindent\textbf{IMU Applications in Respiratory Monitoring.} The development of wearable technologies for monitoring respiratory parameters has seen significant advancements, as demonstrated by several recent studies. One such study \cite{imu_resp1} proposes a system that detects an athlete's breath rate and analyzes correlations with the respiratory minute volume by using gyroscopes and accelerometers to measure thoracic extensions. This system employs two inertial sensors placed on the chest and back of the athlete to determine thoracic extensions differentially, with calculations conducted on a digital signal processor for accuracy. 

Similarly, another research effort \cite{imu_resp2} focuses on using conventional inertial measurement units to measure breathing rates through quaternions. This system, tested against a spirometer, showed minimal variance in respiratory rates, indicating its potential for clinical surveillance tasks. There is a growing interest in integrating wearable devices for respiration monitoring in daily activities, as highlighted by a study \cite{imu_resp3} introducing a wearable Respiration Monitoring System (WRMS) coupled with a Deep Convolutional Neural Network (RasPara-Net DCNN). This system offers a user-friendly and cost-effective solution for real-time respiration monitoring, achieving robust accuracy in predicting breathing rates across various respiration speeds. The WRMS demonstrates significant potential in providing continuous monitoring, contributing to cardiovascular health diagnostics and respiratory assessments. 

By addressing the challenge of continuous monitoring outside clinical settings, another study \cite{imu_resp4} presents a wearable IMU-based device that uses Bluetooth communication with a smartphone to upload data to a web server, allowing for remote monitoring. The wearable, wireless, modular respiratory device illustrates the feasibility of using IMUs for continuous breathing pattern monitoring, even during dynamic conditions. Two additional studies have explored IMUs for monitoring thoracoabdominal respiratory dynamics. The first developed a wearable system for estimating three-dimensional respiratory displacements, achieving high accuracy for respiratory rate and tidal volume \cite{thoraco1}. The second tracked chest movement and estimated respiration rate with IMU sensors, demonstrating high accuracy and cost-effectiveness compared to commercial monitors \cite{thoraco2}. 

Collectively, these studies contribute to highlight the potential of sensors such as accelerometer and gyroscope to provide accurate, continuous, and remote solutions for both clinical and daily use. Although some of these works have explored sensor placement in the thoracoabdominal area, similar to our research, they do not address or target the wider range of respiratory and cardiac conditions examined in our study. Several studies have examined sensor placement for various use cases and applications. However, only a few have used deep learning techniques to extract meaningful features for predicting respiratory dynamics. None of these studies have applied such techniques in a clinical setting to predict cardiorespiratory conditions, as we do.

\medskip

\noindent\textbf{Respiratory Patterns and Sleep Apnea Detection.} Recent advances in sleep apnea detection using inertial measurement units have demonstrated significant potential for non-invasive, continuous monitoring. An approach employing basic machine learning techniques is presented in \cite{5sleepap}; the authors introduced a smartphone application for the detection of obstructive sleep apnea using accelerometer data. Their data acquisition process required wearing a smartphone placed on an arm band, with a microphone near patients' nose, and incorporating a pulse-oximeter. The authors trained a Support Vector Machine for supervised classification, achieving an accuracy of up to 92.2\%.  

A new study \cite{ml_imu2} presents a machine learning-based system that uses two wireless wearable sensors placed on the chest and abdomen to monitor respiratory behavior accurately. By measuring local circumference changes, this system can classify respiratory patterns with high accuracy, addressing the need for continuous monitoring of conditions such as sleep apnea, COPD, and asthma. Another approach \cite{ml_imu4} utilizes a meta-learning algorithm based on Long short-term neural networks to infer respiratory flow from wearables embedding Fiber Bragg Grating sensors and IMUs. This system achieves high accuracy in predicting respiratory flow and estimating parameters like rate and volume, even with a reduced number of sensors, demonstrating the feasibility of unobtrusive, subject-specific monitoring. 

Additionally, a wristwatch device with embedded IMUs has been explored for sleep apnea detection. In a study \cite{ml_imu10} involving 122 adults, the device's triaxial acceleration and gyro signals were used to identify respiratory events, achieving a correlation coefficient of 0.84 with the apnea-hypopnea index from polysomnography. This wristwatch device demonstrated 85\% accuracy in detecting moderate-to-severe sleep apnea and 89\% accuracy for severe cases, highlighting the capability of IMU-based wearables for practical sleep apnea screening and monitoring in daily life.

Wearable technologies have been extensively studied for monitoring respiratory behaviors and cardiovascular health. A first study \cite{ml_imu1} resorted to wireless sensors with machine learning algorithms to continuously monitor respiratory behaviors in various postures, achieving high accuracy in classifying these behaviors and tracking disorders such as COPD, asthma, and apnea. Another study \cite{ml_imu3} explores the use of inertial measurement units (IMUs) to estimate heart rate (HR) and respiratory rate (RR) from chest movements, identifying optimal sensor positions and orientations through data analysis from 15 volunteers, resulting in low mean absolute errors for HR and RR. 

While we also use accelerometer data in our study, our approach is notably different due to our specific research objectives and data acquisition protocol. Unlike previous studies, we incorporate data from both accelerometer and gyroscope sensors placed in multiple locations, which enhances the accuracy and reliability of our measurements. Our proposed architecture, detailed in the next section, is tailored to process this combined sensor data within a well-defined clinical protocol. The dual-sensor approach allows us to classify patients with cardiorespiratory diseases affecting rib cage respiratory kinematics more effectively. Ultimately, our solution leverages affordable sensors commonly found in commercial smartphones. This aims to improve the understanding of their reliability and effectiveness for potential clinical diagnostics, and also opens up possibilities for remote self-screening at home.

\section{Method}
\label{sec:method}

We employed IMU sensors from commodity smartphones, including accelerometer and gyroscope signals over time. For the acquisition proper, we defined a protocol prescribing distinct measurement scenarios, also called \textit{scenes}, wherein each represents a data acquisition performed by placing a smartphone, running our app, over a defined body region of the patient. Subsequently, the acquired data are preprocessed and fed to a tailored deep learning architecture, providing an overall prediction amongst healthy (H) and non-healthy (NH) classes, thereby helping establish the need for further clinical assessments of the individual. It is important to emphasize that this prediction serves solely as a preliminary component of a screening process, applicable in low-resource or pandemic emergency settings.

\subsection{Data acquisition protocol}
\label{subsec:data_acquisition}
The objective of this phase consisted in defining a clinical protocol that could be applied both in a clinical setting with the assistance of trained staff (for the acquisition of the training dataset) and at-home by the patient (for self-administration in the context of a population screening). 

Each measurement session is structured into multiple scenes, with each scene consisting of a given sequence of actions; each action can be one of: (1) starting an IMU measurement of a given duration; (2) initiating an audio recording for a specified period; (3) playing predefined instructory audio or videos from the app. Specifically, within each scene, the subject is directed to adopt a specific posture, position the smartphone in a predetermined location (different for each scene) over the chest or abdomen, and perform a few respiratory cycles as instructed, all of which are recorded by the application\footnote{The app is capable of collecting additional measurement, namely readings from a pulse oximeter and images from a Point-of-Care Ultrasound probe device; however, for the purpose of the present study only the IMU signals were considered.}. For each subject enrolled in the study, the following procedures (as illustrated in Fig.~\ref{fig:chestpositions}) were performed in sequential steps, according to our protocol:

\begin{figure}[ht]
    \centering
    \includegraphics[width=\textwidth]{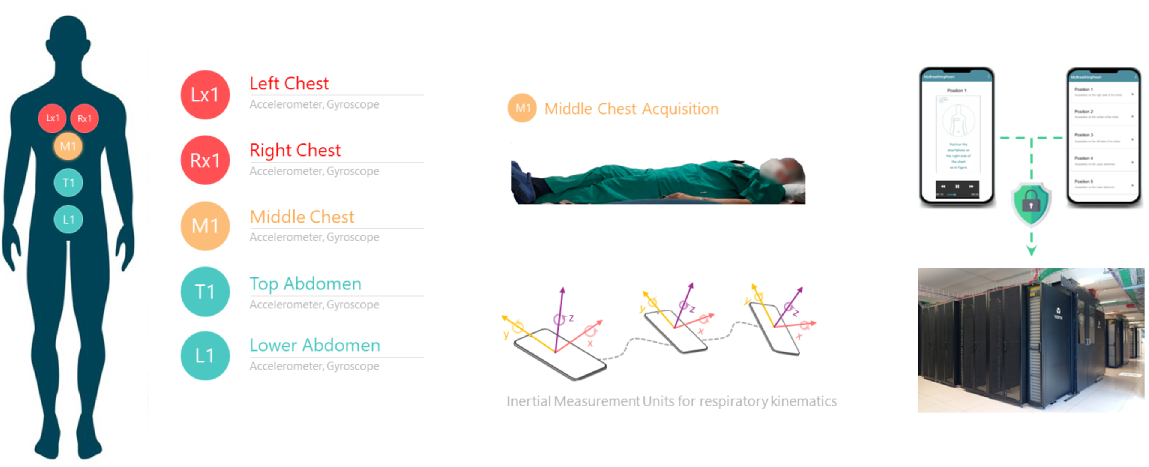}
    \caption{Acquisition protocol. \textbf{(Left)} The five targeted positions for smartphone placement (three on the chest, and two on the abdomen). \textbf{(Middle)} The patient is laying on a bed with a smartphone placed in the middle chest position (M1), illustrating how the smartphone records acceleration and angular velocity through the dedicated app and safely transmits the data to a server \textbf{(Right)}.}
    \label{fig:chestpositions}
\end{figure}

\begin{enumerate}
    \item The patient was instructed to lie supine on a bed.
    \item After ensuring the correct positioning of the subject, the operator placed the smartphone, in one of a set of five predefined configurations, with our app running, where left, right and middle chest are targeted (respectively Lx1, Rx1 and M1), as well as top and lower abdomen (T1, L1).
    \item The patient was then asked to breathe normally for 20 seconds, at the end of which the recording stopped.
    \item Subsequently, the operator moved the smartphone to the next position and repeated the measurement, until all predetermined configurations were completed.
    \item Upon completion of all scenes, the data were securely transmitted to a dedicated server for storage through the app for further processing.
\end{enumerate}
Both the selection of those particular positions and smartphone orientations, and the suitability of typical smartphone IMU sensors, were established in our previous study~\cite{respkine24}. It should be noted that while the above protocol was followed during the initial data acquisition, the entire procedure could also be performed at-home by the patient alone, e.g. as part of a screening program.


\subsection{Dataset and population demographics}
\label{subsec:population}

A population of 133 subjects (77 healthy, 56 preoperative patients) was enrolled when the present study was conducted, relying on an ongoing clinical acquisition and enrollment process for both patients and healthy volunteers. The results presented in this section encompass the entire study population. However, for the experimental results detailed in subsequent sections, it is crucial to note that the reported measurements have undergone quality checks, and specifically refer to data obtained from 77 healthy individuals and 45 patients. Participants were enrolled voluntarily, after individual invitation, and patient volunteers were recruited among those admitted to the hospital of Fondazione Toscana Gabriele Monasterio, presenting a variety of cardiovascular diseases including valvular insufficiency (32), coronary artery disease (17), aortic aneurysm (2). For five patients, the data we received from the medical staff did not include a specific pathology. All participants, including both healthy and patients volunteers, were fully informed about the research purpose of the study before proceeding with the acquisition session. Subjects were presented with the informed consent document, which they conscientiously reviewed and signed. Each subject then underwent an anamnestic questionnaire (see~\ref{appendix:anamnestic}) conducted by the attending physician, to elicit information about their health status profile. The privacy of each subject was ensured by associating measurements to anonymous identifiers. The mapping between identifiers and subjects was maintained separately, outside of our system, according to the relevant regulations.


In this section, we consider the make-up of the population in our study, as pertaining to a subset of selected features (age, sex, weight, height) obtained during the clinical data acquisition phase. We directed our analytical attention to these particular features with an awareness of their potential associations with the signals recorded from the chest and abdominal regions during respiratory cycles. To provide a clearer picture of the patient demographics under investigation, we present in Table~\ref{tab:distribution} the distribution of these features in our population.

\begin{table}[bhtp]
\centering
\begin{tabular*}{\linewidth}{@{\extracolsep{\fill}}llllll}
\toprule
\toprule
\textbf{Label} & \textbf{N} & \textbf{Signals} & \textbf{Age} & \textbf{Height (cm)} & \textbf{Weight (kg)} \\
\midrule
Healthy (H) & 77 & $5 \times 616$ & $42 \pm 11$ & $169.1 \pm 9$ & $66.4 \pm 12.1$ \\
Valvular Insufficiency (NH) & 37 & $5 \times 296$ & $66 \pm 11$ & $169.8 \pm 8$ & $74.8 \pm 13.1$ \\
Coronary Artery Disease (NH) & 17 & $5 \times 136$ & $66 \pm 8$ & $172.2 \pm 7$ & $77.8 \pm 17.0$ \\
Aortic aneurysm (NH) & 2 & $5 \times 16$ & $56 \pm 19$ & $170.0 \pm 10$ & $74.0 \pm 4.0$ \\
\bottomrule
\end{tabular*}
\caption{Summary of subject demographics and signal data across different medical conditions.}
\label{tab:distribution}
\end{table}

\begin{figure}
  \centering
  \hspace*{\fill}
  \subfloat{%
    \includegraphics[width=0.27\textwidth]{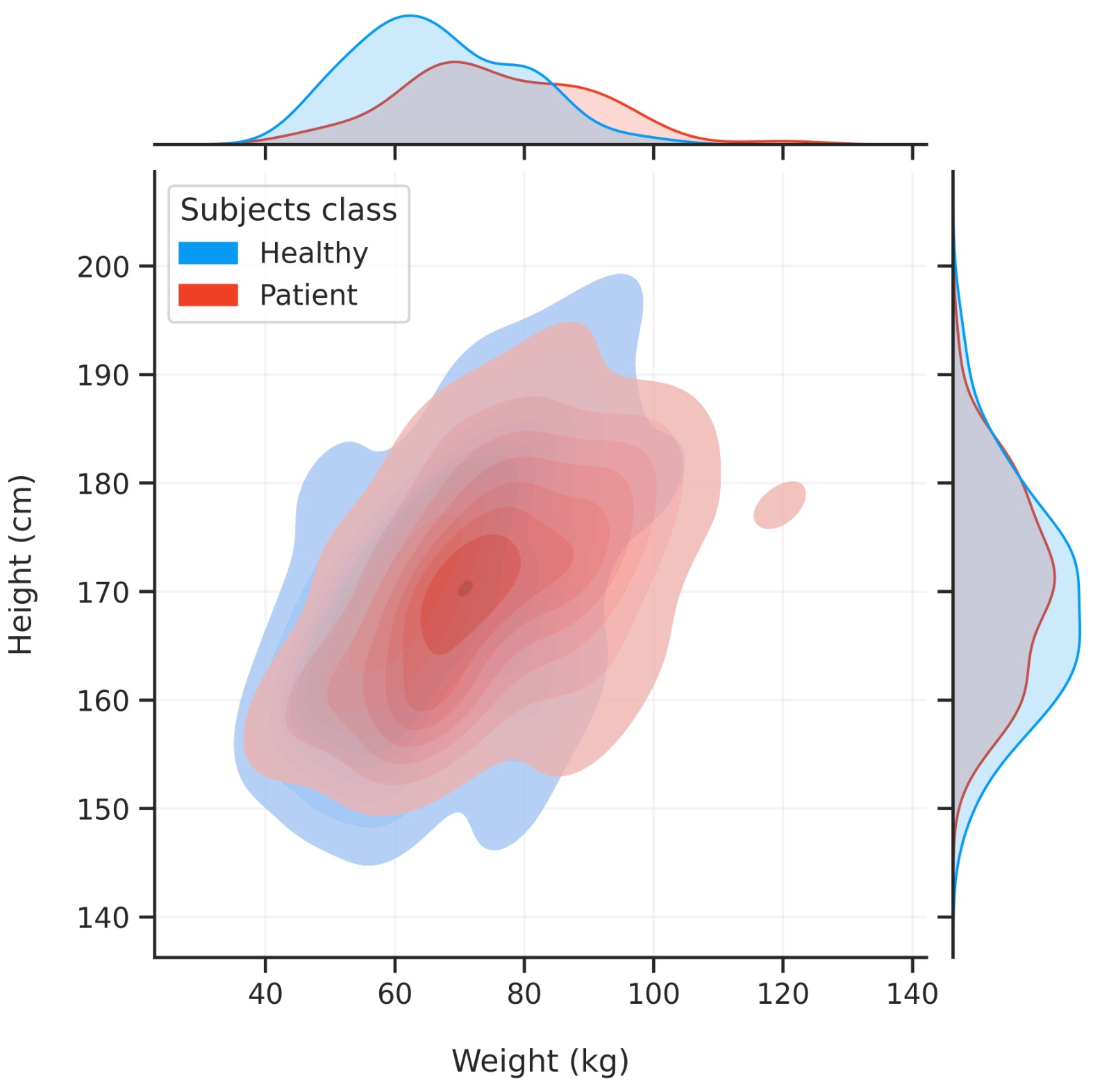}%
    \label{fig:jointplotA}%
  }
  \subfloat{%
    \raisebox{9.3pt}{
    \includegraphics[width=0.7\textwidth]{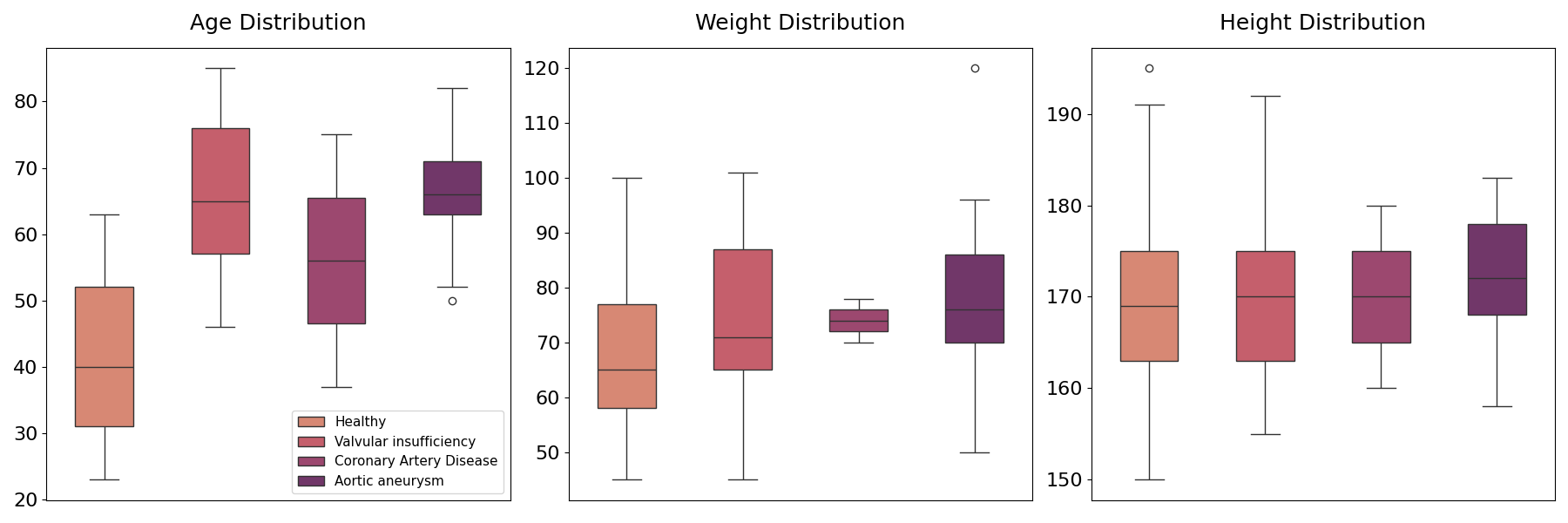}%
    }
    \label{fig:jointplotB}%
  }
  \hspace*{\fill}
  \caption{Joint plot in the first column showing the distribution of height and weight characteristics among healthy and preoperative patients. The remaining columns feature different boxplots illustrating the distribution of age, height, and weight across various health conditions.}
  \label{fig:distribution_plot}
\end{figure}

Within the context of our dataset analysis, we present a joint density plot (Figure \ref{fig:distribution_plot}, first column), which illustrates the correlation between weight and height feature distributions, widely used to compute body mass index (BMI) \cite{bmi}. From the figure, we can distinguish two densities, indicated by red color for patients (NH group) and blue for healthy individuals (H group). There is a clear overlap amongst clusters suggesting that establishing a straightforward correlation between target labels and BMI is not a trivial, and thus biased task. This observation is pivotal in the context of training a predictive model, as the distribution of weight and height features can influence chest movements during breath cycles and measurements through IMU sensors. The remaining three columns in Figure \ref{fig:distribution_plot} presents a boxplot illustrating the distribution of key demographic features across various health conditions. We observed a positive correlation between age and the presence of a generic disease, as patients undergoing surgical procedures were typically older compared to the healthy population. Despite this observed pattern, we conducted an ablation study to assess the influence of age as a predictive feature in our model. We explored both the inclusion of age within input features, as well as training a model separately with this feature. Ultimately, the results consistently showed that our approach effectively captures diagnostically relevant information independent of age-related factors. 

\subsection{Input signals and data preprocessing}
\label{subsec:preprocessing}

\begin{figure}[ht]
    \centering
    \includegraphics[width=\textwidth]{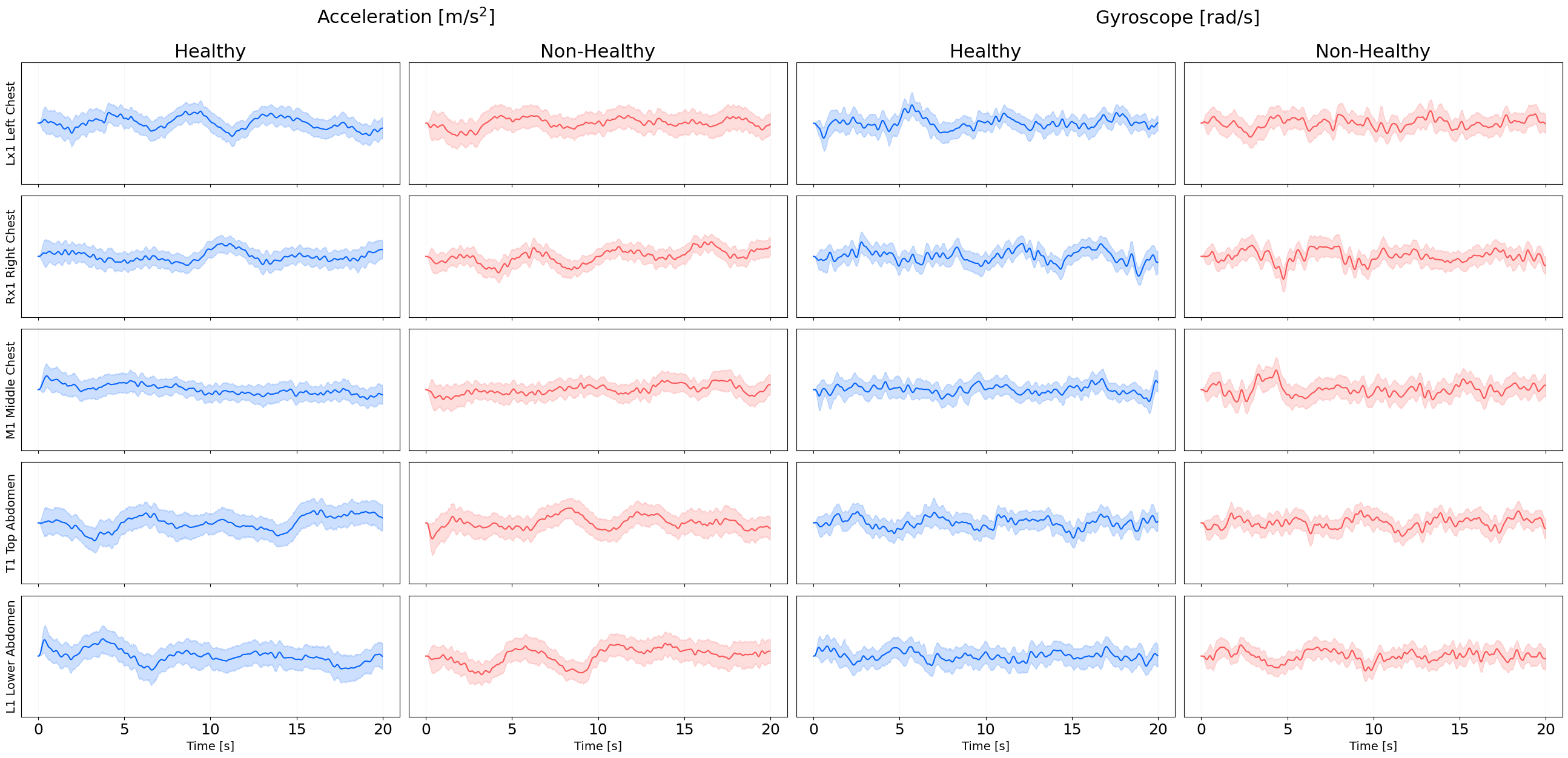}
    \caption{Time series plots of normalized and filtered accelerometer and gyroscope signals from five different recording locations, comparing healthy control and positive class patients. Signals are displayed for a duration of 20 seconds (1000 timesteps) for plotting convenience.}
    \label{fig:full_signal}
\end{figure}

In our clinical context, as we progress through the data acquisition phase, we record a 6-dimensional vector for each of the five previously described scenes, each originating from specific chest or abdomen locations. The resulting input is a multivariate time series, which we will formally denote as $~X_t = [G_t, A_t]$ and comprises data collected from three axes for the accelerometer $A_t = \{(a_{t,x}, a_{t,y}, a_{t,z})\}$ and the gyroscope $G_t = \{(g_{t,x}, g_{t,y}, g_{t,z})\}$, all sampled over a fixed temporal domain $t \in T, 1\leq t \leq m$. The signal $G_t$ represents the gyroscope data at time $t$ providing rate of rotation information through angular velocity measurements along three axes. Similarly, $A_t$ comprises accelerometer measurements capturing changes in velocity influenced by gravitational forces. The time series samples in Figure \ref{fig:full_signal}, represent data from five different sensor placements in each row from both the accelerometer and gyroscope.

\begin{figure}[t]
    \centering
    \subfloat{\includegraphics[width=\textwidth,]{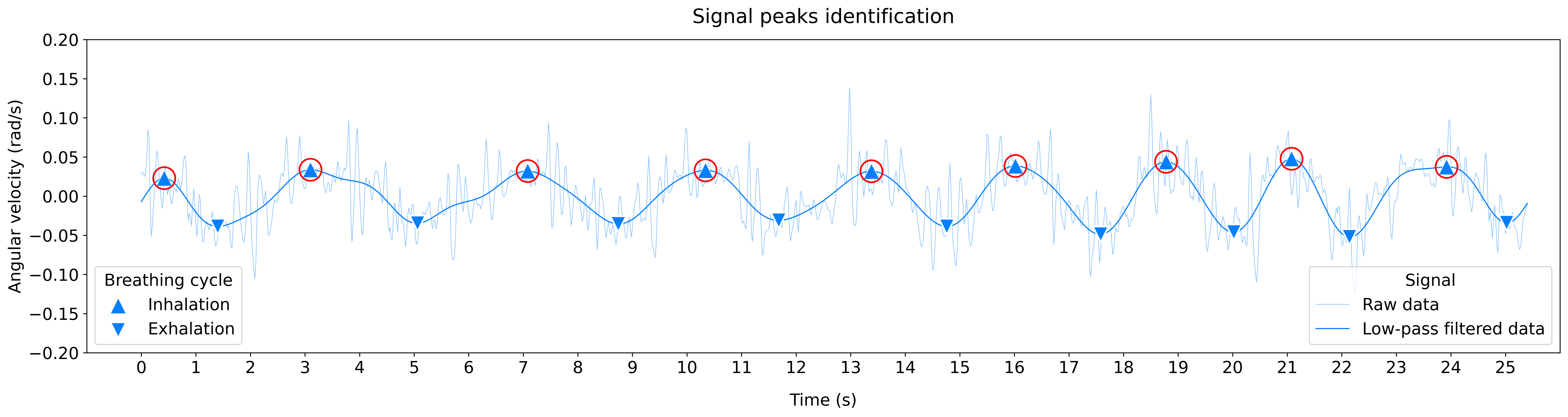}\label{fig:subfig1}} \\
    \subfloat{\includegraphics[width=\textwidth,]{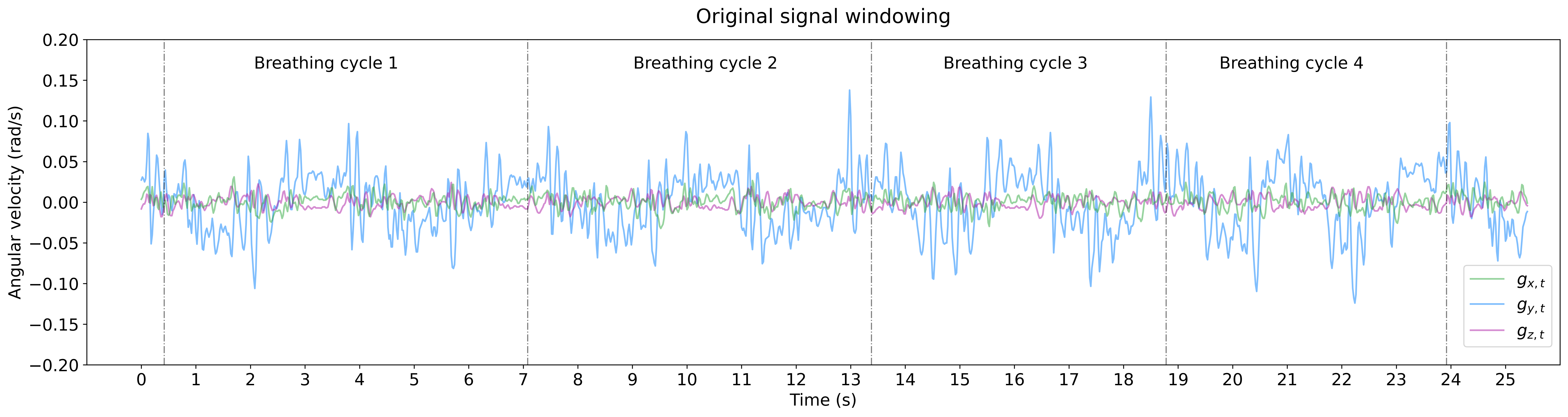}\label{fig:subfig2}}
    \caption{(\textbf{Top}) Low-pass signal filtering on gyroscope Y-axis and peaks identification. Upfacing and downfacing markers denote inhalation and exhalation phases respectively. (\textbf{Bottom}) Signal windowing on original raw data. The same windowing is also applied to time series data coming from the accelerometer.}
    \label{fig:peaks_identification}
\end{figure}

Before proceeding to the analysis of the input signals with deep learning techniques, we performed several preprocessing steps defined in a generalizable pipeline for future acquisitions. The original signal posed multiple challenges such as initial noise recorded from accelerometer and gyroscope readings, attributed to sudden patient movements post smartphone placement. These irregularities were typically attributed to the positioning of the device by medical personnel and involuntary patient movements. Practically, we discarded the first timesteps based on the length of the longest transient observed in $X_t$. This approach was supported by the observation that all transients showed a consistent and short duration (approximately 5 seconds of measurements). Thereafter, we divided the input time series into disjoint shorter sequences, allowing for training data augmentation and localized predictions without suffering from common long-term dependency issues in recurrent neural networks (RNNs). The procedure can be summarized in three crucial steps:

\begin{enumerate}
    \item \textbf{Low-pass filtering.} As a preliminary step, we proceeded to remove noise from the input time series by applying a Fast-Fourier Transform (FFT) based low-pass filter having a cut-off frequency of 0.7Hz. As a result, low-frequency components were highlighted to facilitate the recognition of individual breathing cycles. 
    \item \textbf{Signal peaks identification}. We designed a signal peaks detection procedure by identifying local maxima over the filtered gyroscope y-axis $g_{t,y}$. This choice was supported by empirical observations from the acquisition protocol, which consistently highlighted a major concentration of activity along this axis. The procedure assumes the signal to be smoothed and identifies as a local peak a time step $g_{t,y}$ whenever it has higher amplitude than its direct neighbors $g_{t-1,y}$ and $g_{t+1,y}$. A visual result of these first two steps is depicted in Figure \ref{fig:peaks_identification}~(Top).
    \item \textbf{Original signal windowing}. The original noisy time series $X_t$ is split into single respiratory acts by windowing at the corresponding indexes returned by the previous step, resulting in the identification of breathing cycles as shown in Figure \ref{fig:peaks_identification}~(Bottom). Finally, each of those windows is resampled using FFT transformations for each of the 3D vectors $g_t,a_t$ having a fixed maximum length of $m=300$.
\end{enumerate}

These steps were crucial since they allowed us to isolate individual breathing cycles performed by patients, including a distinction between inhalation and exhalation based on the gyroscope y-axis. The usage of these shorter windows allowed our model to process sequences, which are independent of the actual duration of the measurement session. This approach improved data quality, leading to a more scalable prediction process, and notably increasing the resulting number of training samples. Moreover, by feeding smaller chunks to the model, we drastically reduced training and inference times.

\subsection{Deep Learning architecture}
\label{subsec:dl_architecture}
\begin{figure*}[ht]
\centering
\includegraphics[width=\textwidth,keepaspectratio]{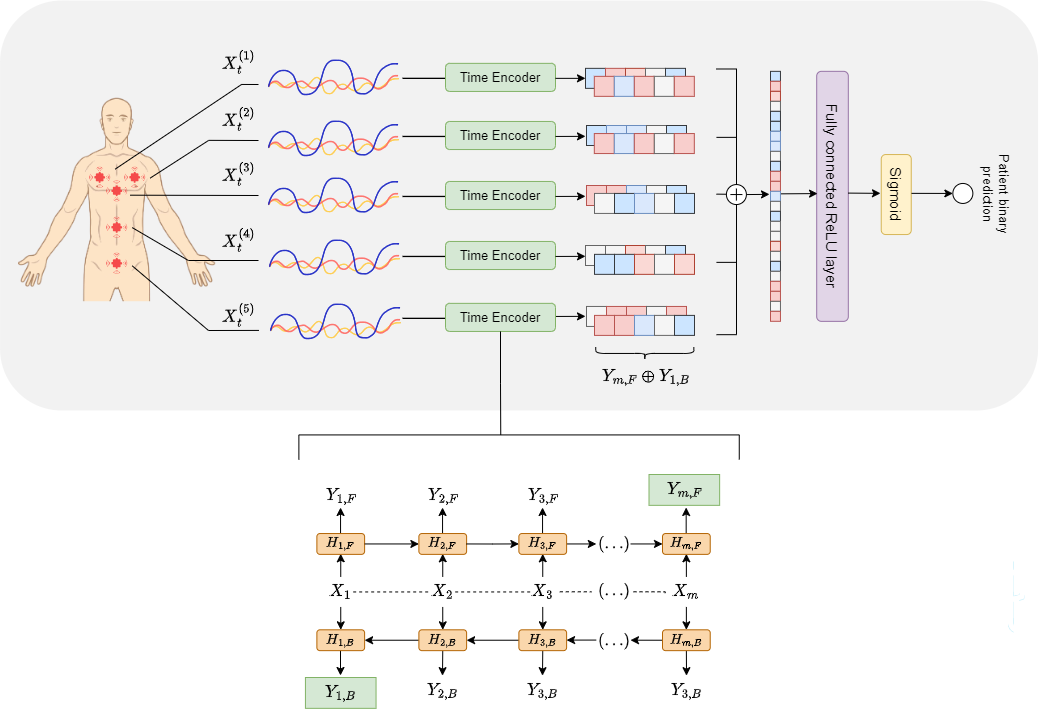}
\caption{Visual workflow of the predictive model. Each processed input signal representing accelerometer and gyroscope data is processed through a bidirectional LSTM encoder. The concatenated embeddings are fed to fully connected layers providing a prediction regarding the presence of a cardiorespiratory disease.}
\label{fig:MBH_arc}
\end{figure*}

The resulting dataset consists of a collection of processed signals from five different acquisition regions for each patient. Since these recordings describe patient's body kinematics throughout breathing cycles, we are interested in designing a recurrent deep learning architecture retaining features for each of these five signals. After an exhaustive model selection phase, as detailed in Section~\ref{sec:results}, we chose a bidirectional LSTM due to its superior performance over validation folds. Given the input $X_t$, the learned features are leveraged yielding a probability score $P(c \mid X_t)$ which corresponds to the confidence of prediction for a generic cardiorespiratory disease as a binary supervised classification task. The overall end-to-end workflow starting from the five input signals and retrieving the binary output is depicted in Figure~\ref{fig:MBH_arc}.

Formally, the input data for our task consists of five multivariate time series $~[X_t^{(1)}, \ldots , X_t^{(5)}]$ with a fixed length ($m=300$) and sampling rate ($50\text{Hz}$) representing individual breathing cycles. The main challenge consists in designing a Time Encoder that adeptly preserves and integrates features in accordance with the characteristics of our application domain. As a solution, we included a bidirectional module processing input in both forward and backward directions. This has shown to be beneficial since it allows to retrieve patterns from both the inhalation and exhalation phases of the breathing cycle. Following the original LSTM formulation \cite{lstm}, the forward \(H_{t,F}\) and backward \(H_{t,B}\) hidden state  at each time step \(t\) are computed as:
\begin{align}
H_{t,F} &= \tanh(W_{i}X_t^{(j)} + U_i H_{t-1,F}) \\
H_{t,B} &= \tanh(W_{i}X_t^{(j)} + U_i H_{t+1,B}),
\end{align}

\noindent where $W_i$ and $U_i$ are the input and recurrent weights, while the previous hidden forward and backward states are denoted as $H_{t-1,F} \in \mathbb{R}^{h}$, $H_{t+1,B} \in \mathbb{R}^{h}$ respectively. For each recorded IMU scene $j$, these hidden states are employed to calculate the forward $Y_{t,F}^{(j)}$ and backward $Y_{t,B}^{(j)}$ outputs as:

\begin{align}
Y_{t,F}^{(j)} &= H_{t,F} \odot \sigma(\overrightarrow{f_t} \odot c_{t-1} + \overrightarrow{i}_t \odot \overrightarrow{\tilde{c}_t}) \\
Y_{t,B}^{(j)} &= H_{t,B} \odot \sigma(\overleftarrow{f_t} \odot c_{t+1} + \overleftarrow{i}_t \odot \overleftarrow{\tilde{c}_t}).
\end{align}

The resulting output is computed based on the original element-wise operations within cell's components including the forget gate $f_t$, the previous candidate $\tilde{c}_t$ and cell state $c_t$, as well as the input gate $i_t$. Then, the model provides the embeddings $E_x$ for a given patient from the concatenation of the first backward $Y_{1,B}$ and last forward output $Y_{m,F}$ from each of the recorded five scenes $E_{x} = Y_{m,F}^{(1)} \oplus Y_{1,B}^{(1)} \oplus \ldots \oplus Y_{m,F}^{(5)} \oplus Y_{1,B}^{(5)}$. The latter, are fed to the head module being a stack of fully connected nonlinear layers with an ending sigmoid neuron, yielding the output probability.

\noindent\textbf{Computational resources for training.}
\label{appendix:computational_resources}
For our model selection and assessment, we leverage GPU parallelization for both training and inference. The top-performing time encoder ($\text{LSTM}_{128,2}$) consists of a total of 2,683,041 trainable parameters, and each epoch (224 samples) with a batch size of 16 elements, was completed in about 4 seconds. The whole training pipeline was executed in approximately 48 hours, with a predefined maximum of 300 epochs and an early stopping patience of 15 for 5 trials of Bayesian search. We employ a configuration comprising:
\begin{itemize}
\setlength\itemsep{-4pt}
    \item {\em Graphics processing units}: 4 $\times$ NVIDIA A100 80GB GPUs
    \item {\em Central processing units}: 2 $\times$ AMD EPYC 7413 24-Core CPUs
    \item {\em Memory:} 1056GB RAM
    \item {\em Operating System:} Linux
\end{itemize}

\section{Results}
\label{sec:results}



\begin{figure}[b]
\centering
\includegraphics[width=\textwidth,keepaspectratio]{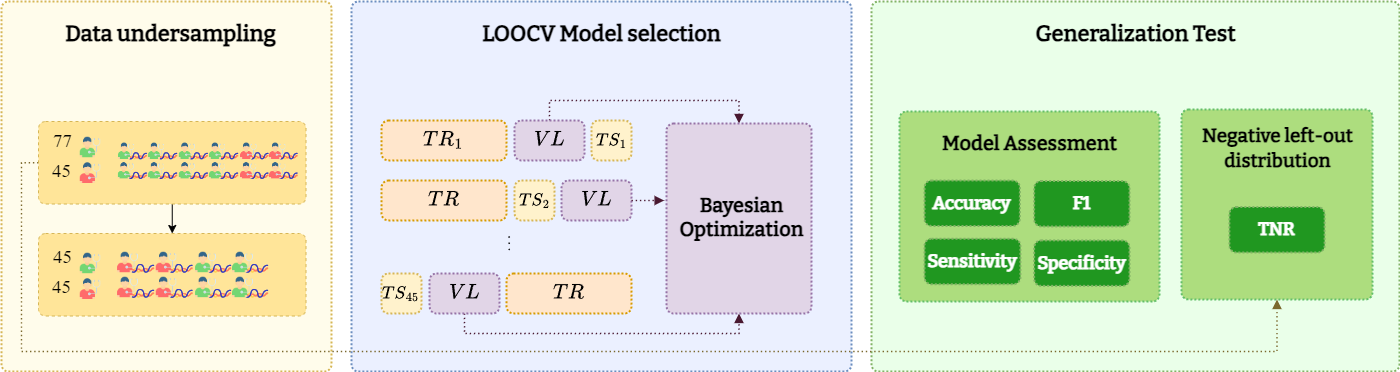}
\caption{Overview of our study's methodology, encompassing dataset undersampling and the application of LOOCV with model assessment. In the left pane, we balance our dataset by uniformly undersampling the majority class. Then, we utilize a LOOCV for robust low-data performance assessment as illustrated in the middle pane, involving iterative testing with singular patients and model selection via Bayesian optimization.}
\label{fig:loocv}
\end{figure}

Before proceeding to the evaluation of various machine learning architectures for the backbone feature encoder, we balanced the dataset to address potential biases by having an objective evaluation of models' performance. We performed a uniformly distributed undersampling procedure over samples from the majority class (healthy patients), repeating the process over four distinct folds representing different seeds. The resulting dataset comprises a total of 90 patients, distributed evenly across the respective classes. In the left pane of Figure \ref{fig:loocv}, we outline the steps of this preliminary procedure. Furthermore, the excluded recordings (32 instances) were not discarded, but rather fully employed in a successive model generalization test over a skewed distribution, thus containing only negative samples. The performance of our models on such a distribution is of particular interest in the context of a general population screening, where a large majority of the subjects would indeed not be affected by the pathologies we have considered.

\subsection{Model selection and assessment}
In the specific context of our application, we opted for a Leave-one-out-cross-validation (LOOCV, \cite{loocv}) as it represents a robust performance evaluation strategy for low-data settings by providing a realistic estimate of model performance on unseen cases. As represented in Figure \ref{fig:loocv} (middle pane), we have established an iterative pipeline wherein, at each iteration step $i$, a singular test patient ($\text{TS}_{\{i\}}$) is chosen, and the remaining samples are partitioned into training (TR) and validation (VL) sets. It is important to point out that this pipeline splits samples based on the patient identifier, guaranteeing complete separation among TR, VL, and test folds during a LOOCV cycle. Within each iteration of the procedure, we performed model selection via Bayesian optimization \cite{snoek2012practical} for hyperparameter tuning. The best model configuration is chosen after training, based on validation performance.

In a concluding evaluation, we assess the generalization capabilities of the selected model, both for the averaged individual test subjects ($\text{TS}_{\{1\}}, \text{TS}_{\{2\}}, \ldots, \text{TS}_{\{n\}} $) and the original set of left-out healthy subjects from the undersampling phase. This enables us to assess not only the model's predictions on unseen cases but also its ability to accurately identify true negatives within a biased distribution, which differs significantly from the balanced one used during training. In the context of medical applications, this is valuable considering that the deployed application in production needs to operate effectively even in scenarios that mirror the real-world distribution of diseases worldwide.

In this study, we conducted a comprehensive investigation, encompassing a diverse range of recurrent deep learning architectures tailored for multivariate time series analysis. We meticulously compared these architectures during an initial model selection phase, which featured state-of-the-art models including Convolutional Neural Networks (CNNs), traditional, bi-LSTMs, and Transformers \cite{transformers}. The performance of each of these architectures as time feature encoders is evaluated based on commonly used metrics (sensitivity, specificity, accuracy) and ranked by F1 metric over validation folds, as shown in Table \ref{tab:val_test_results}. 

\begin{table}[h]
\centering
\resizebox{1\textwidth}{!}{%
\begin{tabular*}{1.1\linewidth}{@{\extracolsep{\fill}}lllllllll@{}}
\toprule
\midrule
 & \multicolumn{2}{c}{Sensitivity} & \multicolumn{2}{c}{Specificity} & \multicolumn{2}{c}{F1} & \multicolumn{2}{c}{Accuracy (\%)} \\
\cmidrule(lr){2-3} \cmidrule(lr){4-5} \cmidrule(lr){6-7} \cmidrule(lr){8-9}
& \small{Validation} & \small{Test} & \small{Validation} & \small{Test} & \small{Validation} & \small{Test} & \small{Validation} & \small{Test} \\
\midrule
$\mathbf{\text{\textbf{Bi-LSTM}}_{128,2}}$ & $\mathbf{0.91_{\pm0.02}}$ & $\mathbf{0.81_{\pm0.02}}$ & $\mathbf{0.93_{\pm0.01}}$ & $\mathbf{0.82_{\pm0.05}}$ & $\mathbf{0.91_{\pm0.02}}$ & $\mathbf{0.81_{\pm0.02}}$ & $\mathbf{90.8_{\pm1.9}}$ & $\mathbf{80.2_{\pm3.9}}$ \\
$\text{Bi-LSTM}_{64,4}$ & $0.81_{\pm0.03}$ & $0.79_{\pm0.05}$ & $0.84_{\pm0.03}$ & $0.81_{\pm0.04}$ & $0.81_{\pm0.03}$ & $0.80_{\pm0.04}$ & $81.5_{\pm3.7}$ & $80.1_{\pm4.1}$ \\
$\text{Bi-LSTM}_{64,2}$ & $0.78_{\pm0.02}$ & $0.78_{\pm0.05}$ & $0.84_{\pm0.02}$ & $0.81_{\pm0.04}$ & $0.79_{\pm0.02}$ & $0.80_{\pm0.04}$ & $78.9_{\pm2.7}$ & $80.1_{\pm4.1}$ \\
$\text{Bi-LSTM}_{32,4}$ & $0.78_{\pm0.03}$ & $0.76_{\pm0.05}$ & $0.82_{\pm0.03}$ & $0.78_{\pm0.04}$ & $0.78_{\pm0.03}$ & $0.77_{\pm0.05}$ & $78.6_{\pm3.3}$ & $76.8_{\pm4.5}$ \\
$\text{LSTM}_{64,4}$ & $0.76_{\pm0.02}$ & $0.76_{\pm0.02}$ & $0.81_{\pm0.02}$ & $0.81_{\pm0.05}$ & $0.77_{\pm0.02}$ & $0.78_{\pm0.02}$ & $77.5_{\pm2.3}$ & $78.8_{\pm3.7}$ \\
$\text{Transformer}_{\text{Deep}}$ & $0.75_{\pm0.02}$ & $0.71_{\pm0.01}$ & $0.79_{\pm0.02}$ & $0.75_{\pm0.04}$ & $0.76_{\pm0.01}$ & $0.73_{\pm0.01}$ & $75.8_{\pm1.8}$ & $73.6_{\pm2.4}$ \\
$\text{LSTM}_{32,6}$ & $0.74_{\pm0.02}$ & $0.70_{\pm0.05}$ & $0.79_{\pm0.02}$ & $0.76_{\pm0.06}$ & $0.73_{\pm0.02}$ & $0.72_{\pm0.06}$ & $74.6_{\pm2.5}$ & $72.3_{\pm5.3}$ \\
$\text{CNN}_{64, 6}$ & $0.71_{\pm0.03}$ & $0.63_{\pm0.03}$ & $0.73_{\pm0.03}$ & $0.63_{\pm0.02}$ & $0.71_{\pm0.02}$ & $0.65_{\pm0.03}$ & $68.7_{\pm2.5}$ & $63.2_{\pm2.4}$ \\
$\text{LSTM}_{32,2}$ & $0.71_{\pm0.01}$ & $0.70_{\pm0.01}$ & $0.77_{\pm0.01}$ & $0.76_{\pm0.03}$ & $0.70_{\pm0.02}$ & $0.72_{\pm0.01}$ & $71.7_{\pm1.4}$ & $72.5_{\pm1.5}$ \\
$\text{Transformer}_{\text{Base}}$ & $0.67_{\pm0.01}$ & $0.67_{\pm0.01}$ & $0.71_{\pm0.0}$ & $0.63_{\pm0.02}$ & $0.67_{\pm0.01}$ & $0.61_{\pm0.02}$ & $67.7_{\pm0.9}$ & $63.6_{\pm1.3}$ \\
$\text{CNN}_{128, 6}$ & $0.52_{\pm0.03}$ & $0.50_{\pm0.01}$ & $0.63_{\pm0.02}$ & $0.53_{\pm0.03}$ & $0.61_{\pm0.01}$ & $0.57_{\pm0.02}$ & $54.4_{\pm0.6}$ & $53.3_{\pm2.0}$ \\
$\text{CNN}_{128, 12}$ & $0.53_{\pm0.04}$ & $0.53_{\pm0.03}$ & $0.62_{\pm0.03}$ & $0.56_{\pm0.03}$ & $0.58_{\pm0.02}$ & $0.57_{\pm0.05}$ & $52.9_{\pm1.2}$ & $49.6_{\pm3.6}$ \\
\bottomrule
\end{tabular*}
}
\caption{Performance evaluation of various time encoder architectures for cardiorespiratory disease prediction using accelerometer and gyroscope signals. The table presents results ranked in decreasing order based on the F1 metric over validation folds. Additionally, the same metrics are shown for the test set.}
\label{tab:val_test_results}
\end{table} 

The model employing a bi-LSTM (two stacked LSTM layers with 128 units each) as a feature extractor, consistently outperformed its counterparts across all evaluated metrics. Interestingly, the performance trends observed in the validation set extends also to the averaged test set results, and the inability of 1D CNNs to compete across various kernel sizes and filters highlights the complexity of the designated task. Given the sequential nature of the input breathing cycles characterized by distinct phases of inhalation and exhalation, this architecture played a crucial role; its ability to simultaneously analyze the input sequence in both directions allowed the model to grasp useful patterns throughout various phases of the respiratory cycle. 

Transformers and CNNs with varying complexities showed comparatively lower sensitivity and specificity. The conventional transformer architecture faced several challenges due to its inability to capture sequential dependencies, making it difficult to discern patterns within time series data. Additionally, a global focus during self-attention may hinder their effectiveness in prioritizing relevant features, especially in tasks where specific time windows hold significant importance, such as cardiorespiratory disease detection.

\begin{figure}[tbhp]
    \centering
    \includegraphics[width=\textwidth, keepaspectratio]{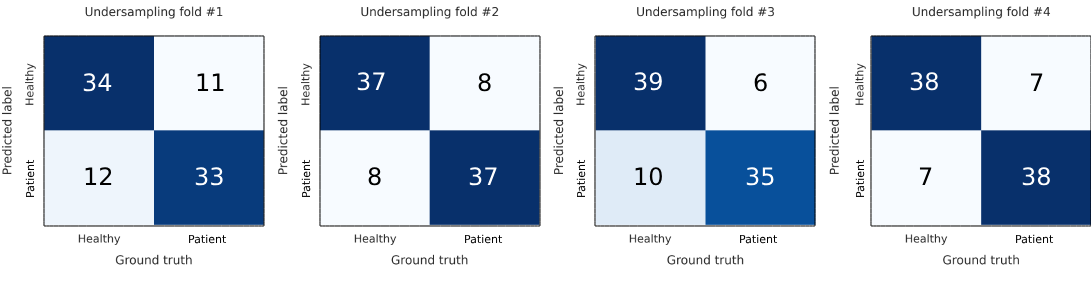}
    \caption{The confusion matrix shows the number of true positives (TP), true negatives (TN), false positives (FP), and false negatives (FN) for each undersampled fold. Target labels are shown along the y-axis columns, and model predictions are shown along the x-axis rows.}
    \label{fig:test_cm}
\end{figure}

Distinct confusion matrices corresponding to unique random seeds for the exclusion of healthy subjects during the preliminary undersampling stage are shown in Figure \ref{fig:test_cm}. The model exhibits consistent performance across diverse undersampling folds, refuting the hypothesis of any bias in test results favoring specific subsets of subjects characterized by notable disparities between positive and negative classes.


\subsection{Diseases and breathing cycle analysis}

Due to the complexity of performing multi-classification given the limited amount of singular cases for each disease, we disposed them into macro-categories or classes and assigned them corresponding labels. This allows us to further specialize the accuracy results obtained for our binary case. The objective is to uncover potential trends in disease recognition to assess whether the model exhibited variations in performance across disease classes. In Table \ref{tab:disease_acc} we report the performance for the best architecture evaluated over the various classes of pathologies. The overall reported accuracy results imply a degree of similarity in the model's performance across various disease classes, without a distinct leader among the evaluated cardiovascular conditions. This uniformity promotes the potential of the methodology to tackle varied and complex classification tasks within the cardiorespiratory domain, leveraging recordings and methodologies of this nature.

\begin{table}[ht]
\centering
\begin{tabular}{lccc}
\toprule
\midrule
Class & Accuracy & Misclassified & N \\
\midrule
Healthy & $82.2_{\pm 4.2}$ & (11,8,6,7) & 45\\
Coronary Artery Disease & $81.3_{\pm3.6}$  & $(3,2,2,2)$ & 12 \\
Valvular Insufficiency & $77.9_{\pm6.9}$  & $(8,4,7,4)$ & 26\\
Aortic aneurysm & $100.0_{\pm 0.0}$ & $(0,0,0,0)$ & 2\\
\bottomrule
\end{tabular}

\caption{Summary of performance metrics for each of the reported diseases on the test set. The values in parentheses depict the distribution across different classes, obtained from four different seeds in the initial undersampling procedure across various patients. The table provides accuracy percentages, misclassified instances, correct instances, and the total number of instances (N) for each labeled disease.}
\label{tab:disease_acc}
\end{table}

\begin{figure}[t]
\centering
\includegraphics[width=1\textwidth,keepaspectratio]{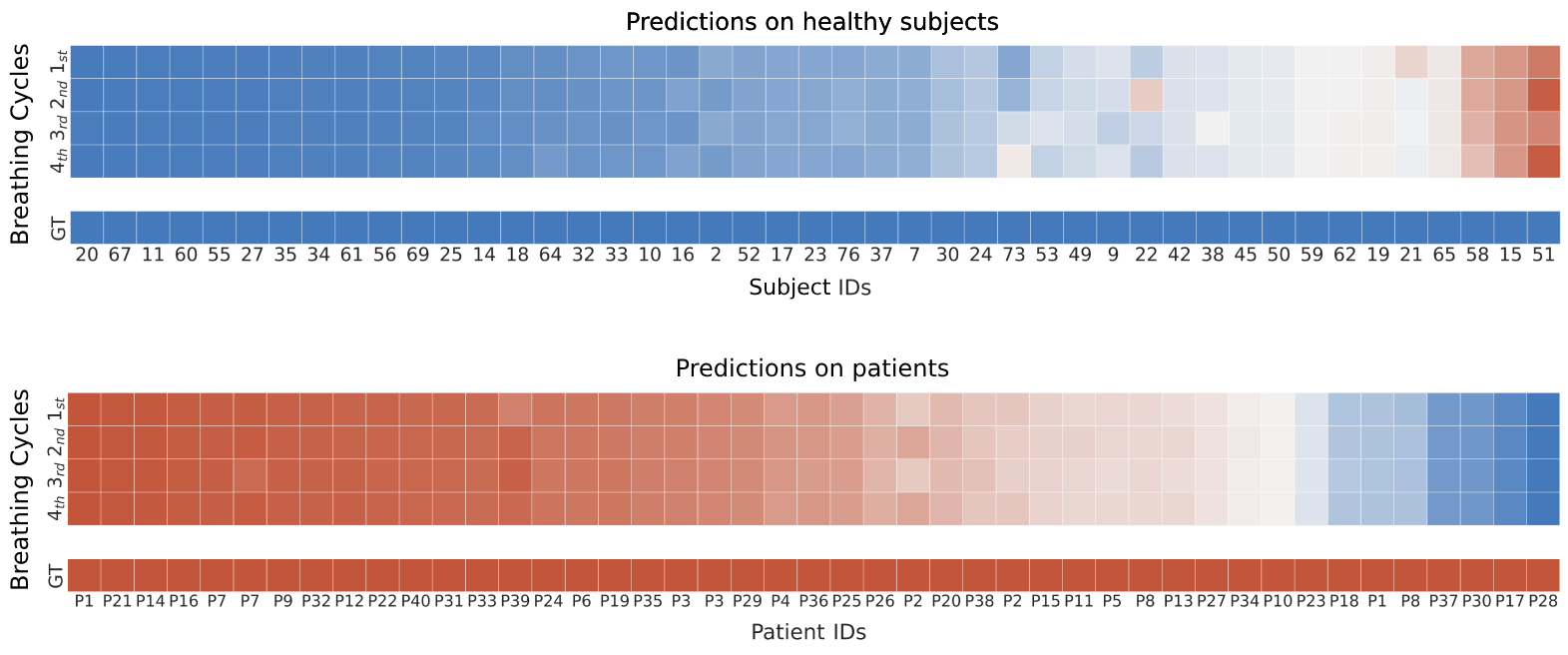}
\caption{Heatmap representation of predictions for positive (top heatmap) and negative (bottom heatmap) patients across different test splits. Each heatmap consists of four rows, illustrating predictions for individual breathing cycles, with the detached row at the bottom indicating the ground truth (GT) values. The color spectrum transitions from blue (denoting negative predictions with a value of 0) to red (representing positive predictions with a value of 1) as predicted by the model.}
\label{fig:heatmap}
\end{figure}

In Figure \ref{fig:heatmap}, we illustrate a heatmap representation of predictions for subjects categorized as healthy (H group, top plot) and patients (NH group, bottom plot) across individual test splits in LOOCV. Each heatmap is organized into four rows, symbolizing predictions for the first four individual breathing cycles of a measurement, while a detached row below represents ground truth (GT) values. The color spectrum smoothly transitions from blue, representing predictions with a value of $0$ as negative cases, to red, indicating predictions valued at $1$ as positives. This visual representation offers a comprehensive view of the model's predictive outcomes. The latter remains steady and consistent across the breathing cycles for the majority of the patients in these folds. Moreover, the observed low variability in predictions over breathing cycles from the same patient contributes to emphasizing the effectiveness of the preprocessing stages.

\section{Discussion}
\label{sec:discussion}
\subsection{Clinical application}

Although this study focuses on the enrollment of pre-operative patients with various cardiorespiratory diseases, the underlying objective of our cheap non-invasive tool extends beyond this specific cohort, aiming to address clinical needs in the evolving era of remote digital health. The application's inherent affordability and remote capabilities position it as a low-cost decentralized solution, avoiding the need for on-site technicians. This characteristic facilitates its implementation in diverse clinical settings, encompassing remote and domiciliary environments where direct medical intervention is not readily available.

The cost-effectiveness and remote capabilities of our solution render it well-suited for the preliminary screening of large patient populations at risk, particularly relevant in pandemic contexts, such as the recent COVID-19 outbreak. Notably, the affordability of the solution permits individual evaluations at home, utilizing commonly available smartphones. The latter is also an opportunity to collect extensive data for refining and optimizing the underlying solution, serving as a foundational model for subsequent investigations and improvements. Moreover, the proposed solution promotes the development of low-cost hardware, incorporating inertial sensors potentially more cost-effective than smartphones. Thus, holding the premises for further applications in resource-constrained clinical settings, which would expand the reach of these technologies as a screening tool.

The solution is also easily applicable to the monitoring of diseases that affect chest kinematics during breathing, such as those observed in post-operative or chronic patients \cite{k1,k2,k3,k4}. The system's ability to monitor changes in predictions or features over time presents the prospect of interpreting disease progression, thus allowing for more precise and individualized assessments of specific cases. In summary, the technology's capacity for cost-effective remote screening, combined with its potential for data-driven refinement and adaptability to various clinical conditions, positions it as a promising tool for improving patient care across a range of medical contexts.

\subsection{Limitations}
\nointitem{Long-Term Monitoring and Follow-Up:}
While our study demonstrates promising short-term predictive accuracy in predicting cardiorespiratory diseases through respiratory kinematics via IMU sensors on common smartphones, it is important to acknowledge the limitation in capturing the model's performance over extended periods. The focus on short-term outcomes may not fully reflect the model's effectiveness in long-term monitoring and follow-up of patients with cardiorespiratory diseases. This aspect warrants further investigation and consideration for practical clinical implementation, where continuous and sustained monitoring is often crucial for managing chronic conditions.

\nointitem{Limited Data Size and Diversity:}
An inherent limitation of the study lies in the dataset's size and diversity. The restricted volume of data may not comprehensively cover the entire spectrum of cardiorespiratory diseases, potentially influencing the model's generalizability. The scarcity of data for certain conditions may lead to biased predictions if applied to populations or diseases that are underrepresented in our dataset.

\nointitem{Heterogeneity of Patient Population:}
The study's findings are subject to the inherent heterogeneity in patient demographics and health conditions present within the dataset. This diversity poses challenges in ensuring the model's robustness across varied age groups and condition-specific differences. The effectiveness of the predictive model may be influenced by unaccounted variations in patient characteristics, emphasizing the need for the ongoing enrollment and acquisition of diverse patients coming from different clinical contexts.

\nointitem{Sensor Precision and Placement:}
The accuracy of IMU sensors on commodity smartphones, and of their placement on the chest and abdomen, introduce potential limitations in capturing nuanced respiratory kinematics. Variations in precision and placement could impact the model's predictive accuracy, leading to challenges in consistently and accurately capturing subtle changes. Further refinement in the preprocessing stage and acquisition protocols has been of substantial importance in addressing these limitations.

\nointitem{Sensitivity to Smartphone Models:}
Our study acknowledges the sensitivity of the predictive model to variations in smartphone models and their built-in sensors. Inconsistencies in data acquisition arising from different smartphone specifications may impact the model's performance. Recognizing this limitation, efforts should be directed toward standardizing data collection protocols or ensuring model adaptability to diverse smartphone technologies for reliable and consistent predictions.

\nointitem{External Factors and Noise:}
Environmental factors and external noise during data collection, including ambient conditions and user movement, present potential sources of interference that may introduce noise and affect prediction accuracy. Understanding the impact of these external factors is essential for interpreting the model's performance under real-world conditions. Future research should explore additional strategies to mitigate noise and enhance the robustness of predictions in challenging environments, contributing to improved reliability in clinical applications.

\section{Conclusions}
\label{sec:conclusions}
In summary, this research introduces an innovative method leveraging commodity smartphones' IMU sensors and deep learning techniques for the early and large-population screening of cardiorespiratory diseases. The incorporation of Leave-one-out-cross-validation with Bayesian optimization for hyperparameter tuning and model selection adds a rigorous and data-driven dimension to the evaluation process. The proposed deep learning model, having as a backbone a bidirectional LSTM, exhibited high accuracy as well as sensitivity, specificity, and F1 score across diverse test folds. The application of our technology extends beyond pre-operative patients, positioning itself as a low-cost decentralized solution with remote capabilities. Its affordability and suitability for diverse clinical settings make it a promising tool for preliminary screening, especially in pandemic contexts like the recent COVID-19 outbreak. The adaptability of the solution for monitoring diseases affecting global and local respiratory kinematics, such as post-operative or chronic conditions, further underscores its potential for individualized assessments and disease progression monitoring. Additionally, the research underscores the potential of widely available smartphones as effective tools for timely cardiorespiratory disease screening in diverse clinical settings, offering crucial assistance to public health efforts. The combination of cost-effectiveness, remote capabilities, and adaptability positions this technology as a promising asset for improving patient care with low-cost devices and relatively moderate training resources.

\section*{Author contributions}

Conceptualization: LS, LMi, VG, SC; Methodology: LS, LMi, VG; Software: LS, LMi, VG, LMo; Validation: LS, LMi, VG; Formal analysis: LS, LMi; Investigation: LS, LMi, LMo, EV, EG, SC; Resources: EV, EG, SC; Data curation: VG, EV, EG; Writing - original draft: LS, LMi, VG; Writing - review \& editing: LS, LMi, VG, EV, EG, SC; Visualization: LS, LMi, VG, EV, EG, SC; Supervision: VG, SC; Project administration: SC; Funding acquisition: VG, SC.
All authors have read and agreed to the published version of the manuscript.

\section*{Funding}
The research was funded by the Tuscany Regional Government, in part under the ``Progetti speciali COVID-19'' program, project ``MyBreathingHeart: Sviluppo ed implementazione di un’applicazione per smartphone per il monitoraggio remoto di problemi cardio-respiratori durante crisi pandemica'', CUP I55F20000610007, and in part under the ``Assegni di ricerca 2021'' program, project ``VoBiLaI4H: Vocal Biomarkers \& Language Intelligence for Health'', CUP I53D21002340008.

\section*{Ethics}
The Regional Ethics Committee for Clinical Experimentation of Tuscany approved this study (code 20417\_BERTI). Written informed consent was obtained from all participants in the study. The authors declare no conflicts of interests.

\bibliographystyle{elsarticle-num} 
\bibliography{refs}

\appendix

\section{Anamnestic questionnaire}
\label{appendix:anamnestic}

\begin{table}[H]
  \centering
  \renewcommand{\arraystretch}{1.5} 
  \small
  \begin{tabular}{|p{0.06\linewidth}|p{0.66\linewidth}|P{0.15\linewidth}|}
    \hline
    \textbf{QID} & \textbf{Question} & \textbf{Answer} \\
    \hline
    
    \multicolumn{3}{|c|}{\textbf{1. Demographic Information}} \\
    \hline
    Q.1.1 & Age &  \\ \hline
    Q.1.2 & Sex &  \small\fbox{M} \small\fbox{F} \\ \hline
    Q.1.3 & Weight &  \\ \hline
    Q.1.4 & Height &  \\ \hline
    
    \multicolumn{3}{|c|}{\textbf{2. General health information}} \\
    \hline
    Q.2.1 & Smoking status & \small\fbox{Y} \small\fbox{N} \small\fbox{Former} \\ \hline
    Q.2.2 & Pregnancy & \small\fbox{Y} \small\fbox{N} \\ \hline
    Q.2.3 & Passive exposure to respiratory substances & \small\fbox{Y} \small\fbox{N}  \\ \hline
    
    \multicolumn{3}{|c|}{\textbf{3. Medical History}} \\
    \hline
    Q.3.1\textsuperscript{*} & Allergies & \small\fbox{Y} \small\fbox{N} \\ \hline
    Q.3.2\textsuperscript{*} & Chronic pharmacological treatment & \small\fbox{Y} \small\fbox{N}  \\ \hline
    Q.3.3\textsuperscript{*} & Chronic respiratory diseases (without pharmacological treatment) & \small\fbox{Y} \small\fbox{N}  \\ \hline
    Q.3.4\textsuperscript{*} & Chronic respiratory diseases (with pharmacological treatment) & \small\fbox{Y} \small\fbox{N}  \\ \hline
    Q.3.5\textsuperscript{*} & Autoimmune diseases & \small\fbox{Y} \small\fbox{N}  \\ \hline
    
    \multicolumn{3}{|c|}{\textbf{4. Recent Health Status}} \\
    \hline
    Q.4.1\textsuperscript{*} & Fever or other flu-like symptoms (last week) & \small\fbox{Y} \small\fbox{N}  \\
    \hline
    Q.4.2 & Past infection with SARS-CoV-1 & \small\fbox{Y} \small\fbox{N}  \\ \hline
  \end{tabular}
  \caption{Anamnestic questionnaire provided to patients before the study. The superscript (\textsuperscript{*}) indicates that if the answer is yes, the patient has to complement it with additional information related to the question.}
  \label{table:patient_info}
\end{table}

\end{document}